\documentclass{article}

\PassOptionsToPackage{square,numbers}{natbib}
\usepackage[final]{neurips_2023}

\usepackage[utf8]{inputenc} 
\usepackage[T1]{fontenc}    
\usepackage{hyperref}       
\usepackage{url}            
\usepackage{booktabs}       
\usepackage{amsfonts}       
\usepackage{nicefrac}       
\usepackage{microtype}      
\usepackage{xcolor}         
\usepackage{hyperref}

\usepackage{geometry}
\usepackage{tabularx}
\usepackage{graphicx}
\usepackage{multirow}
\usepackage{afterpage}

\title{Exploring Generalisability of Self-Distillation with No Labels for SAR-Based Vegetation Prediction}

\author{%
  Laura Martínez-Ferrer \\
  Universitat de Val\`encia, Spain\\
  \texttt{laura.martinez-ferrer@uv.es}
  \And
  Anna Jungbluth \\
  European Space Agency, Climate Office, UK\\
  \texttt{anna.jungbluth@esa.int} 
  \And
  Joseph A. Gallego-Mejia \\
  Universidad Nacional de Colombia, Colombia\\
  \texttt{jagallegom@unal.edu.co} 
  \And
  Matt Allen \\
  University of Cambridge, UK\\
  \texttt{mja78@cam.ac.uk} 
  \And
  Francisco Dorr \\
  Independent, Argentina\\
  \texttt{fran.dorr@gmail.com} 
  \And
  Freddie Kalaitzis \\
  University of Oxford, UK\\
  \texttt{freddie.kalaitzis@cs.ox.ac.uk} 
  \And
  Raúl Ramos-Pollán \\
  Universidad de Antioquia, Colombia\\
  \texttt{raul.ramos@udea.edu.co}
}

\begin{document}

\maketitle

\begin{abstract}
In this work we pre-train a DINO-ViT based model using two Synthetic Aperture Radar datasets (S1GRD or GSSIC) across three regions (China, Conus, Europe). We fine-tune the models on smaller labeled datasets to predict vegetation percentage, and empirically study the connection between the embedding space of the models and their ability to generalize across diverse geographic regions and to unseen data. For S1GRD, embedding spaces of different regions are clearly separated, while GSSIC's overlaps. Positional patterns remain during fine-tuning, and greater distances in embeddings often result in higher errors for unfamiliar regions. With this, our work increases our understanding of generalizability for self-supervised models applied to remote sensing. 
\end{abstract}


\section{Introduction}
Earth observation (EO) and remote sensing techniques have witnessed significant advancements in recent years, allowing us to characterize our planet with ever increasing resolution and accuracy. Using EO data, we can e.g. characterize global vegetation in a consistent manner, and link vegetation changes to the complex feedback mechanisms of climate change\citep{ustin2021current}. Synthetic Aperture Radar (SAR) data \citep{LIU2019506} has proven immensely valuable for this task, since SAR sensors are sensitive to the dielectric and geometric characteristics of plants, and provide consistent, all-weather, day-and-night monitoring capabilities. However, making sense of the wealth of (unlabelled) SAR data requires advanced computational techniques such as machine learning (ML) and specifically self-supervised learning (SSL) which has revolutionized the training of neural networks without the need for labeled data. Self-supervised learning with SAR images is a new research direction and has not been widely explored \citep{9875399}, and can therefore greatly enhance the scalability and cost-effectiveness of SAR-based vegetation estimation \citep{alzubaidi2023survey}.

In this study, we investigate the application of self-\textbf{di}stillation with \textbf{no} labels (DINO)\citep{caron2021emerging} to SAR-based vegetation estimation. We specifically focus on understanding how the embedding space of the SSL pre-trained model empirically links to generalizability across diverse geographic regions (in our case Europe, China, the Continental United States (Conus), and South America), and when fine-tuning with limited labelled data. By assessing model performances on areas outside of the training set, we uncover valuable insights into the models' robustness and scalability. More specifically, we show that shared embedding spaces across geographic regions improve the generalizability of pre-trained models to previously unseen regions. With this, our work significantly advances our understanding of the applicability and reliability of SSL techniques for remote sensing, with profound implications for environmental science and land management, where labelled datasets might be limited.

\section{Data}

\textbf{Areas of Interest \& Data Splits.  }
Our dataset includes four regions of interest (AOIs): Europe, China, the Continental United States (Conus), and South America. For each region, we created ML-ready datasets of SAR images and associated vegetation percentage. Among the AOIs, SSL pre-training was performed on three areas (Europe, Conus, and China), while downstream fine-tuning was performed on each AOI separately. 
We partitioned all data into tiles measuring 448$\times$448 pixels, using \texttt{geetiles}\footnote{\url{https://github.com/rramosp/geetiles}} and \texttt{sartiles}\footnote{\url{https://github.com/rramosp/sartiles}} for pre-processing. These tiles were further divided into training (60\%), validation (20\%), and test (20\%) sets based on geographic bands to minimize data leakage across contiguous tiles (see Figure \ref{fig:data-splits} in the SI). To keep the data volume manageable, we only used data from 2020 in this work.

\textbf{SAR Data.  }\label{sec:input}
We processed two data diversities from \href{https://sentinels.copernicus.eu/web/sentinel/missions/sentinel-1}{Sentinel-1}: (1) SAR coherence from the \href{https://asf.alaska.edu/data-sets/derived-data-sets/sentinel-1-interferograms/}{Global Seasonal Sentinel-1 Interferometric Coherence and Backscatter} (GSSIC) dataset \cite{Kellndorfer2022} and (2) SAR amplitude from the \href{https://developers.google.com/earth-engine/datasets/catalog/COPERNICUS_S1_GRD}{Sentinel-1 Level 1 Ground Range Detected} (S1GRD) dataset. While the amplitude characterizes the intensity of the back-scattered signal in vertical (V) or horizontal (H) polarisation, coherence measures the similarity of two complex SAR images taken at different times, i.e. using both intensity and phase information.
For GSSIC, we used seasonal averages (spring, summer, autumn, winter) for all possible 12 and 24 timedeltas available to calculate SAR coherence using VV\footnote{I.e. For pulses sent and received in vertical polarization.} polarization, totalling 8 channels. For S1GRD we used seasonal averages for polarizations VV and VH\footnote{I.e. For pulses sent in vertical and received in horizontal polarization.} including the logarithmic difference between them (VV-VH), resulting in 12 channels. 

\textbf{Vegetation Data.  }
After pre-training on unlabeled SAR data, we fine-tuned our models for vegetation estimation, using the \href{https://developers.google.com/earth-engine/datasets/catalog/MODIS\_006\_MOD44B}{Terra MODIS Vegetation Continuous Fields} (VCF) product. For this, we calculated the yearly mean vegetation percentage per SAR tile as labels for the supervised fine-tuning. The distribution of labels for each AOI is shown in Figure \ref{fig:hist} in the SI.

\section{Self-Supervised Learning \& Fine-tuning using DINO}
\begin{figure}[ht]
  \centering
  \includegraphics[trim=0 220 20 0, clip,width=\textwidth]{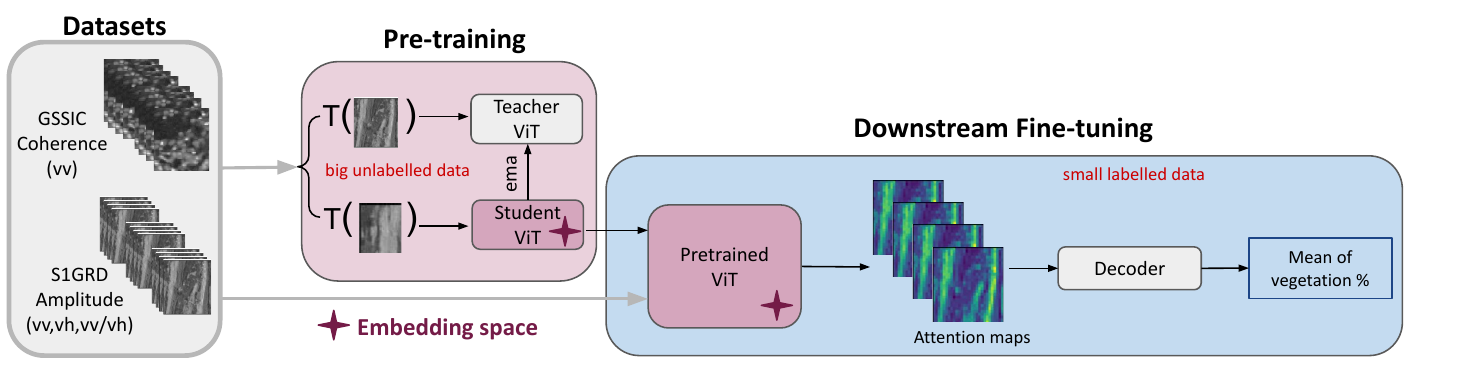}
  \caption{Diagram of the ML pipeline used in this work. GSSIC (8 channels) or S1GRD (12 channels) SAR data is used as an unlabelled input to DINO. After pre-training, the attention maps generated by the student backbone are used as input to a linear decoder to predict mean vegetation percentage. The pink stars show the steps in the pipelines that were used to visualize model embeddings.}
  \label{workflow}
\end{figure}

Self-\textbf{di}stillation with \textbf{no} labels (DINO) \citep{caron2021emerging} is an SSL method based on a student-teacher framework, where the student network aims to emulate the representations generated by the teacher. While the teacher network has access to the full input, the student only sees cropped components of the image to learn context from limited information. The weights of the teacher are updated using an exponential moving average (ema) from the student, and the student and teacher embeddings are matched using a cross-entropy loss.
In this work, we adapted the original DINO architecture for SAR data. We used ViTs as the student and teacher backbones to make use of the ViTs attention mechanism for model fine-tuning (see Figure \ref{fig:input-attention-maps} in the SI). Due to SAR specifics, we only employed horizontal flipping and cropping as transformations during pre-training.
Figure \ref{workflow} shows a diagram of our ML pipeline. We use either GSSIC or S1GRD as unlabeled input to DINO during pre-training. Afterwards, we use the attention maps generated by the pre-trained student backbone as the input to a linear decoder to predict the mean vegetation percentage for each SAR tile. While we also performed experiments where the weights of the backbone are frozen, the best results were obtained when both the backbone and the decoder were trained during fine-tuning. To rationalize model performance and investigate generalizability, we analyzed the embeddings of the pre-trained or fine-tuned backbone (labelled with pink stars in Figure \ref{workflow}).

 
\section{Results}

\textbf{Benefits of Model Pre-training. }
For both S1GRD or GSSIC, our model was pre-trained in China, Conus, and Europe, totalling $\sim$650K input tiles. After pre-training, the model was fine-tuned on each AOI separately. 
Table \ref{tab:scratch_rmse} in the SI shows the root mean square error (RMSE[\%]) in predicting mean vegetation percentage when training the model from scratch or using the pre-trained backbone. 
In all cases, supervised fine-tuning was performed using only 1\% of available data in each AOI, to investigate the performance of SSL when labelled datasets are limited. With more data, the RMSE[\%] reduces slightly (see Figure \ref{fig:hist} in the SI). We generally observe lower RMSE[\%] with pre-training, and therefore focus the remaining discussion on the pre-trained models.

\begin{figure}[tb]
    \centering
    \includegraphics[width = 0.8\linewidth]{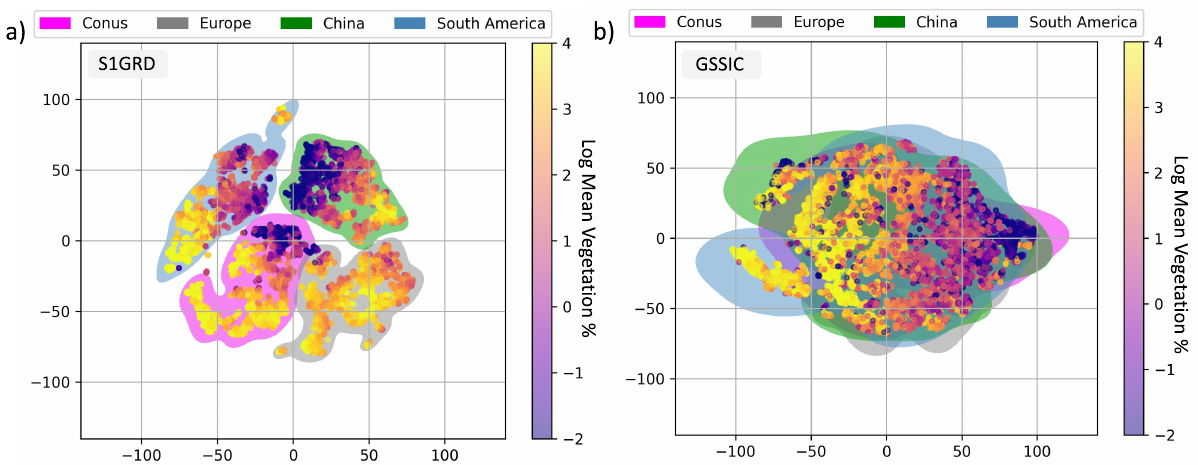}
    \caption{Comparison of the embedding space of the pre-trained DINO model for a) S1GRD, and b) GSSIC, for tiles from Conus (pink), Europe (grey), China (green), and South America (blue). The embedding space was visualized using the class token and via TSNE dimensionality reduction. For each area, we show the embedding of 2000 tiles from the test set (dots), colored with each tile's mean vegetation percentage.}
    \label{fig:pretraining-embeddings}
\end{figure}

\textbf{Empirical Investigation of Embedding Spaces \& Model Generalizability. }
The generalizability of SSL models hinges on meaningful embedding spaces, which should capture essential and distinguishing data features for effective generalization to new, unseen data. This requires thoughtful design of pretext tasks and model architectures. Figure \ref{fig:pretraining-embeddings} shows a 2D TSNE reduction of the embeddings from each AOI after pre-training DINO using S1GRD (\ref{fig:pretraining-embeddings}a) or GSSIC (\ref{fig:pretraining-embeddings}b). The positional embedding of each tile is colored with the logarithm of the mean vegetation percentage. Importantly, the vegetation labels are not used to compute the embeddings or the 2D TSNE position, and are completely unseen by the model at this stage. 
The embeddings of the pre-trained models show clear positional patterns, and similar values of vegetation are clustered together. In the case of S1GRD, all AOIs are clearly separated, while the embedding space of GSSIC overlaps regions in a continuous label distribution. 
We hypothesize that the complexity of features contained in the different SAR input data contributes to the observed differences. S1GRD utilizes amplitude data from multiple polarizations, while our GSSIC input relies only on VV polarization coherence. In high density vegetation regions, the forest structure sensitivity of coherence leads to low values \citep{GHOSH2021104737}, this may hinder the model's coherence-vegetation relationship learning when predicting it. In addition, the S1GRD dataset has higher resolution (448x448 pixels/tile compared to $\sim$ 50x60 pixels/tile for GSSIC) and it provides more information about local features. Therefore S1GRD likely has more data diversity than GSSIC, resulting in lower RMSE[\%] values (see Table \ref{tab:scratch_rmse} in SI). 

To empirically assess generalizability, we visualize the embedding space after model fine-tuning and investigate whether model performance on previously unseen regions can be predicted from looking at the embedding space alone.
More specifically, we plot the 2D TSNE reduction of the embedding space after fine-tuning on 1\% of data in Europe, Conus, or China, and investigate how well vegetation in South America (which has not been seen) is predicted. In Figure \ref{fig:embedding_south} the first (third) column shows the embeddings and prediction error for 2000 example tiles in the respective fine-tuning datasets for S1GRD (GSSIC), while the second (fourth) column shows the embeddings and prediction errors when performing inference of the fine-tuned models on 2000 tiles in South America. Similar plots for inference on the training regions are included in the Supplementary Information, but not discussed here.
Looking at Figure \ref{fig:embedding_south}, we observe that after fine-tuning using S1GRD, the embedding spaces of the different AOIs remain separated. Generally, prediction errors are high (yellow dots), for tiles that are far from the embedding space of the fine-tuning AOI. We quantified embedding separation using Sliced Wasserstein Distance (SWD), which is a mathematical metric used for comparing high-dimensional distributions \cite{kolouri2019generalized}. All SWD values were calculated for 10 seeds and 10000 projections. The lowest SWD and RMSE[\%] for predicting vegetation in South America is observed when using the model fine-tuned on Conus. For the model fine-tuned on Europe or China, both the SWD and RMSE[\%] values are high, although no direct correlation between them was observed.
In the case of GSSIC, the embedding spaces of all AOIs still show large overlaps after fine-tuning. We empirically observe that prediction errors in South America match those in the fine-tuning AOIs for shared embeddings. In other words, good predictions in the model fine-tuned on e.g. Conus translate to low errors for South America in the same embedding space. We observe that prediction errors increase if tiles in South America fall further outside the embedding space of the fine-tuning AOI. For Europe-fine-tuned model, South America's embedding space barely overlaps with the few tiles in Europe that the model is able to predict well, resulting in the highest RMSE for South American inference. For both input modalities, the embedding space of Europe shows the largest SWD to the embedding space of South America. Looking at the distribution of vegetation across the AOIs, we hypothesize that the stark difference in vegetation found in Europe compared to the other regions leads to the poor prediction performance when applying the model fine-tuned in Europe to other regions.


\section{Conclusions}
To summarize, we pre-trained DINO on unlabelled SAR data, and later fine-tuned our models on small datasets across diverse geographic regions to predict mean vegetation percentage. We find that pre-training leads to lower errors compared to training from scratch. This highlights the potential of SSL for remote sensing applications, where labeled datasets are limited. Furthermore, we visualized the embedding spaces of the pre-trained and fine-tuned models to draw conclusions about their generalizability. We empirically find that the complexity of the data could lead to larger distances in embedding spaces (measured via the SWD) and therefore cause larger prediction errors for unseen regions as is the case of S1GRD dataset. However, it is important to acknowledge the limitations of this work, and further research is needed to analyze additional datasets. The exploration should involve finding a metric to characterize the complexity and diversity of the data, thereby strengthening the support for our hypothesis. With this, our work presents a first step towards creating fully generalizable models for Earth Observation.
\afterpage{%
\begin{figure}[ht]
    \centering
    \includegraphics[width=0.98\linewidth]{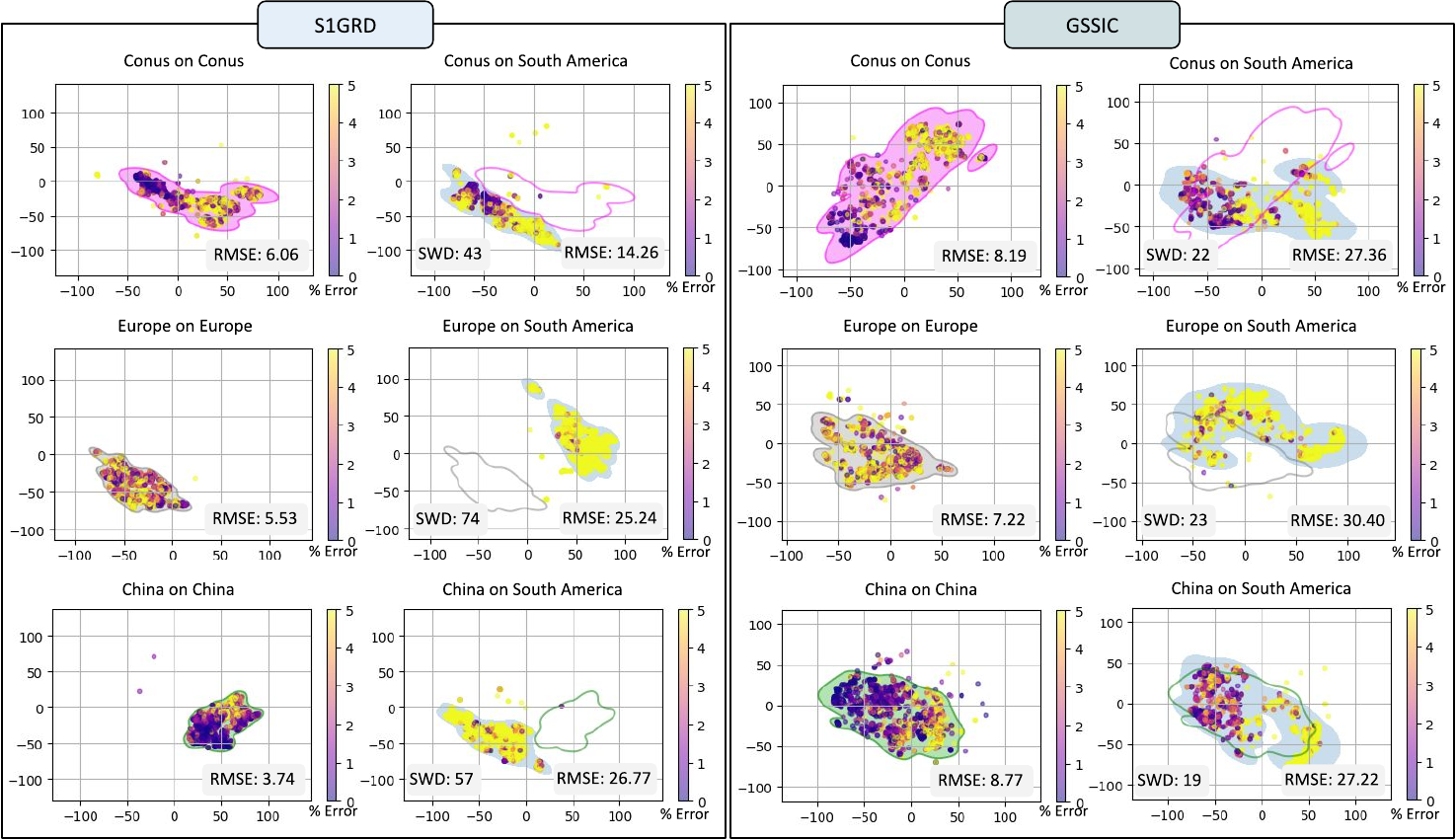}
    \caption{Embedding space and prediction errors (in \%) for models pre-trained on S1GRD or GSSIC. The first (third) column shows the embeddings and prediction error for 2000 example tiles in the respective fine-tuning areas for S1GRD (GSSIC), while the second (fourth) column shows the embeddings and prediction errors when performing inference of the fine-tuned models on 2000 tiles in South America. The Sliced Wasserstein Distance (SWD) of the embedding distributions and the root mean square error (RMSE[\%]) for inference on the entire test is included.}
    \label{fig:embedding_south}
\end{figure}
\section{Acknowledgements}
This work has been enabled by Frontier Development Lab Europe (\url{https://fdleurope.org}) a public / private partnership between the European Space Agency (ESA), Trillium Technologies, the University of Oxford and leaders in commercial AI supported by Google Cloud and Nvidia, developing open science for all Humankind.  L.M-F. was supported by the European Research Council (ERC) Synergy Grant “Understanding and Modelling the Earth System with Machine Learning (USMILE)” under the Horizon 2020 research and innovation programme (Grant agreement No. 855187). M. J. A. was supported by the UKRI Centre for Doctoral Training in Application of Artificial Intelligence to the study of Environmental Risks [EP/S022961/1], and additionally by Trinity Hall, Cambridge. We are also indebted to Nicolas Longépé, Carlos López-Martínez, Fabio A. González Osorio, Samuel Bancroft, Emma Hatton, Alison Lowndes, Alistair Francis, Ioanna Bouri and the rest of reviewers during 2023 FDL-Europe sprint.
}
\afterpage{%
\clearpage
\bibliographystyle{unsrt}
\bibliography{refs}
}
\afterpage{%
\newpage
\appendix
\section{Supplementary Information}

\begin{figure}[h]
    \centering
    \includegraphics[trim=0 80 20 0, clip,width = 0.9\linewidth]{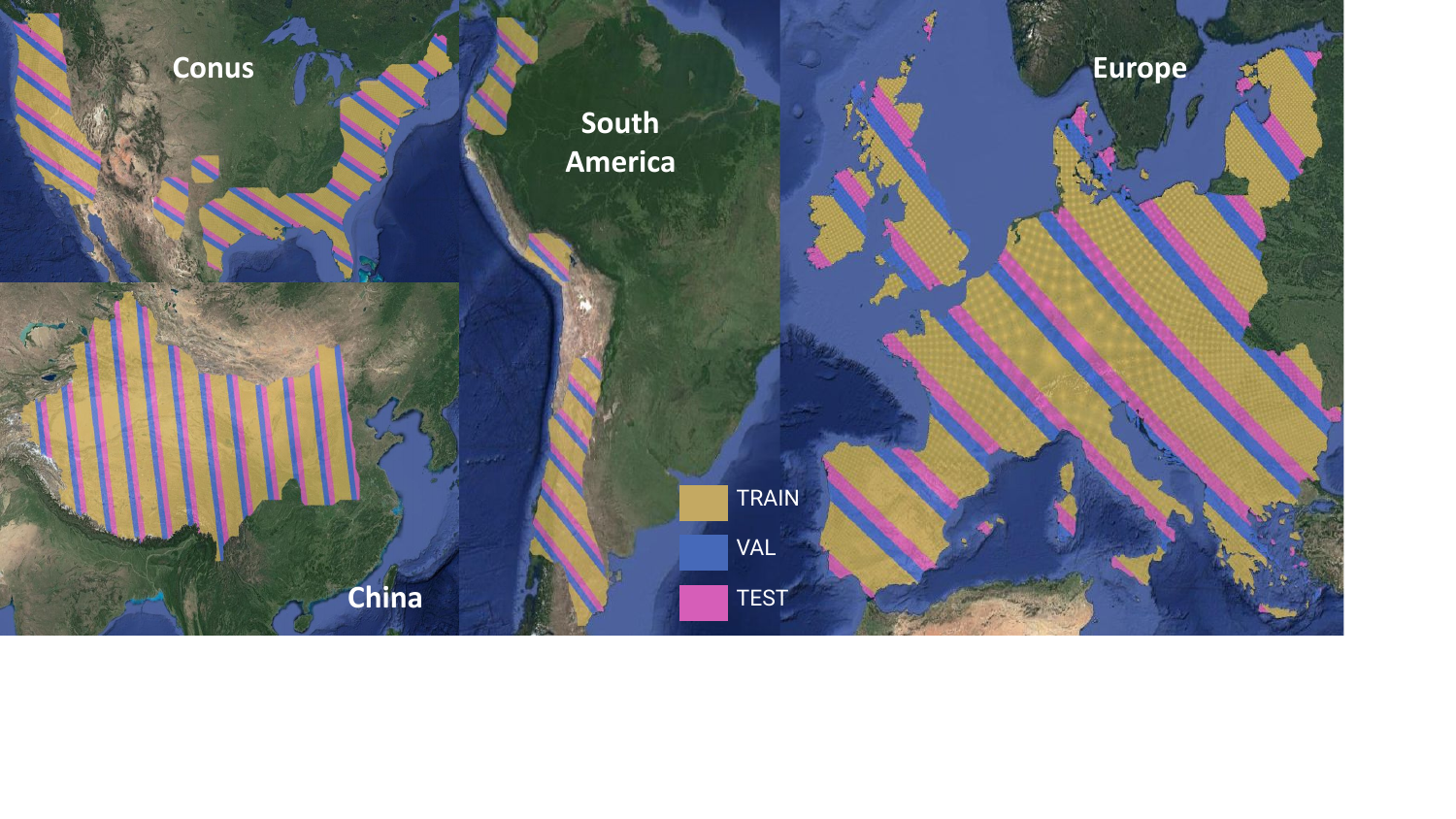}
    \caption{Data splits for Europe, Conus, China, and South America. In total there are 167K image tiles for Conus, 285K tiles for China, 200K tiles for Europe, and 83K tiles for South America.}
    \label{fig:data-splits}
\end{figure}

\begin{table}[h]
    \caption{Parameters used for pre-training DINO on unlabelled S1GRD or GSSIC datasets.}
    \centering
    \begin{tabular}{c|c|c}
    & GSSIC & S1GRD \\
    \hline
    Model Architecture & ViT\_Base & ViT\_Base \\
    Learning Rate & 0.00001 & 0.000001 \\
    Teacher Temperature & 0.04 & 0.001 \\
    Student Temperature & 0.1 & 0.03 \\
    Warm-up Teacher Temperature & 0.04 & 0.01 \\
    Warm-up Teacher Temperature Epochs & 10 & 5 \\
    Center Momentum & 0.90 & 0.99 \\
    \hline
    \end{tabular}
    \label{tab:training_params}
\end{table}

\begin{figure}
    \centering
    \includegraphics[width = \linewidth]{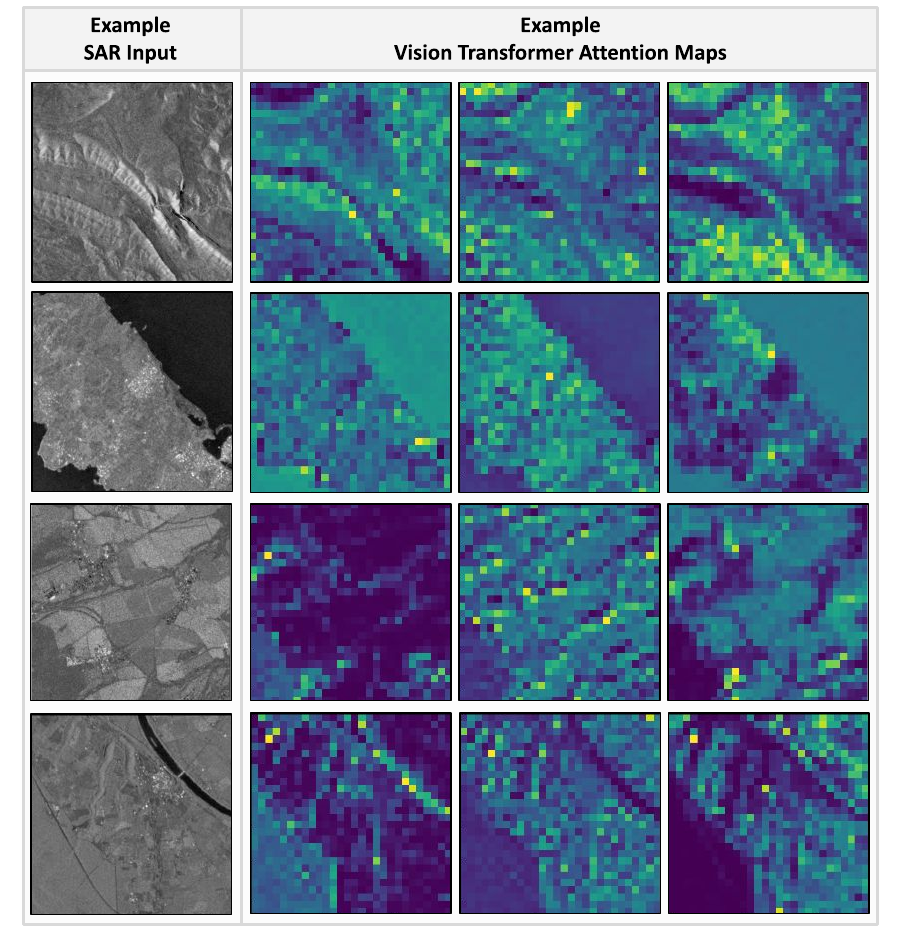}
    \caption{Example S1GRD SAR input images and ViT attention maps. The ViT used during this work have 12 attention heads that pay attention to different features in the input data.}
    \label{fig:input-attention-maps}
\end{figure}

\begin{figure}[ht]
    \centering
    \includegraphics[width=\textwidth]{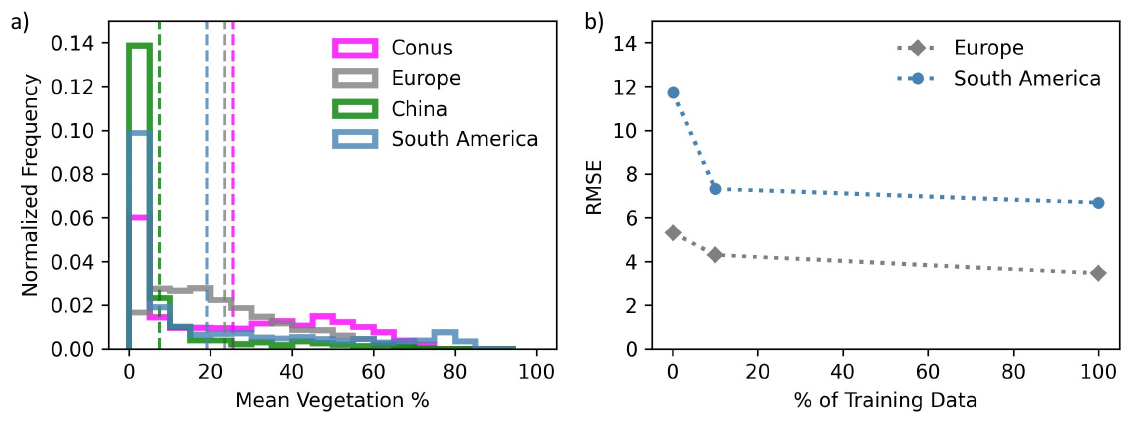}
    \caption{a) Histogram distributions of the mean vegetation percentage labels for each AOI, and b) root mean square error (RMSE[\%]) in predicting mean vegetation percentage after fine-tuning on Europe (grey) or South America (blue) using 1\%, 10\% or 100\% of available data. Analyzing the distributions of vegetation percentage in different AOIs can offer insights into the distribution learned by the model during the fine-tuning process. We observe that for Europe, which was part of the pre-training set, the improvement in RMSE[\%] when fine-tuning on 100\% compared to 1\% of data is very small. Even for South America, which was not part of the pre-training set, good model performances are obtained for small data fine-tuning.}
    \label{fig:hist}
\end{figure}

\begin{table}[ht]
    \centering
    \caption{Root mean square error (RMSE[\%]) for predicting mean vegetation percentage using S1GRD or GSSIC. The model was either trained from scratch or using our pre-trained backbone with 1\% of available data in Europe, China, Conus, or South America.}
    \begin{tabular}{c|c|c|c|c}
        \multicolumn{5}{c}{S1GRD} \\
        \hline
        & Europe & China & Conus & South America \\
        \hline       
        From scratch & 5.50 & 3.39 & 5.58 & \textbf{10.14} \\
        With pre-training & \textbf{5.31} & \textbf{3.12} & \textbf{5.30} & 11.73 \\
        \hline
        \\[-2ex] 
        \multicolumn{5}{c}{GSSIC} \\
        \hline
        & Europe & China & Conus & South America \\
        \hline  
        From scratch & \textbf{6.33} & 7.29 & 8.29 & 11.09 \\
        With pre-training & 6.78 & \textbf{6.81} & \textbf{7.54} & \textbf{10.77} \\        
        \hline
    \end{tabular}
    \label{tab:scratch_rmse}
\end{table}

\begin{table}[ht]
  \caption{Root mean square error (RMSE[\%]) values when fine-tuning (columns) and performing inference (rows) in each AOI. All fine-tuning was done using only 1\% of available data, and all inferencing was performed on the entire test set. The values in \textcolor{purple}{purple} refer to the models trained on S1GRD, and the values in \textcolor{teal}{teal} refer to the models trained on GSSIC. For completeness, we also fine-tuned a model on $\sim$ 1\% of data sampled from the combined area of Europe, Conus, and China, shown in the last row.}
  \centering
  \begin{tabular}{c|c|c|c|c}
    & Conus & Europe & China & South America \\
    \hline
    Conus & \textcolor{purple}{6.06} / \textcolor{teal}{8.19} &\textcolor{purple}{9.53} / \textcolor{teal}{15.56} &  \textcolor{purple}{22.05} / \textcolor{teal}{15.00} & \textcolor{purple}{14.26} / \textcolor{teal}{27.36} \\
    \hline
    Europe & \textcolor{purple}{18.29} / \textcolor{teal}{13.80} &\textcolor{purple}{5.53} / \textcolor{teal}{7.22} &  \textcolor{purple}{20.93} / \textcolor{teal}{19.70} & \textcolor{purple}{25.24} / \textcolor{teal}{30.40} \\
    \hline
    China & \textcolor{purple}{27.93} / \textcolor{teal}{31.90} &\textcolor{purple}{20.71} / \textcolor{teal}{25.34} &  \textcolor{purple}{3.74} / \textcolor{teal}{8.77} & \textcolor{purple}{26.77} / \textcolor{teal}{27.22} \\
    \hline
    South America & \textcolor{purple}{21.12} / \textcolor{teal}{19.88} &\textcolor{purple}{16.36} / \textcolor{teal}{21.10} &  \textcolor{purple}{17.18} / \textcolor{teal}{12.60} & \textcolor{purple}{12.72} / \textcolor{teal}{11.87} \\
    \hline
    Conus/Europe/China & \textcolor{purple}{6.33} / \textcolor{teal}{11.33} &\textcolor{purple}{5.77} / \textcolor{teal}{9.71} &  \textcolor{purple}{4.38} / \textcolor{teal}{10.60} & \textcolor{purple}{18.95} / \textcolor{teal}{14.49} \\
    \hline
  \end{tabular}
  \label{tab:embedding_prediction}
\end{table}

\begin{figure}[ht]
    \centering
    \includegraphics[width=\linewidth]{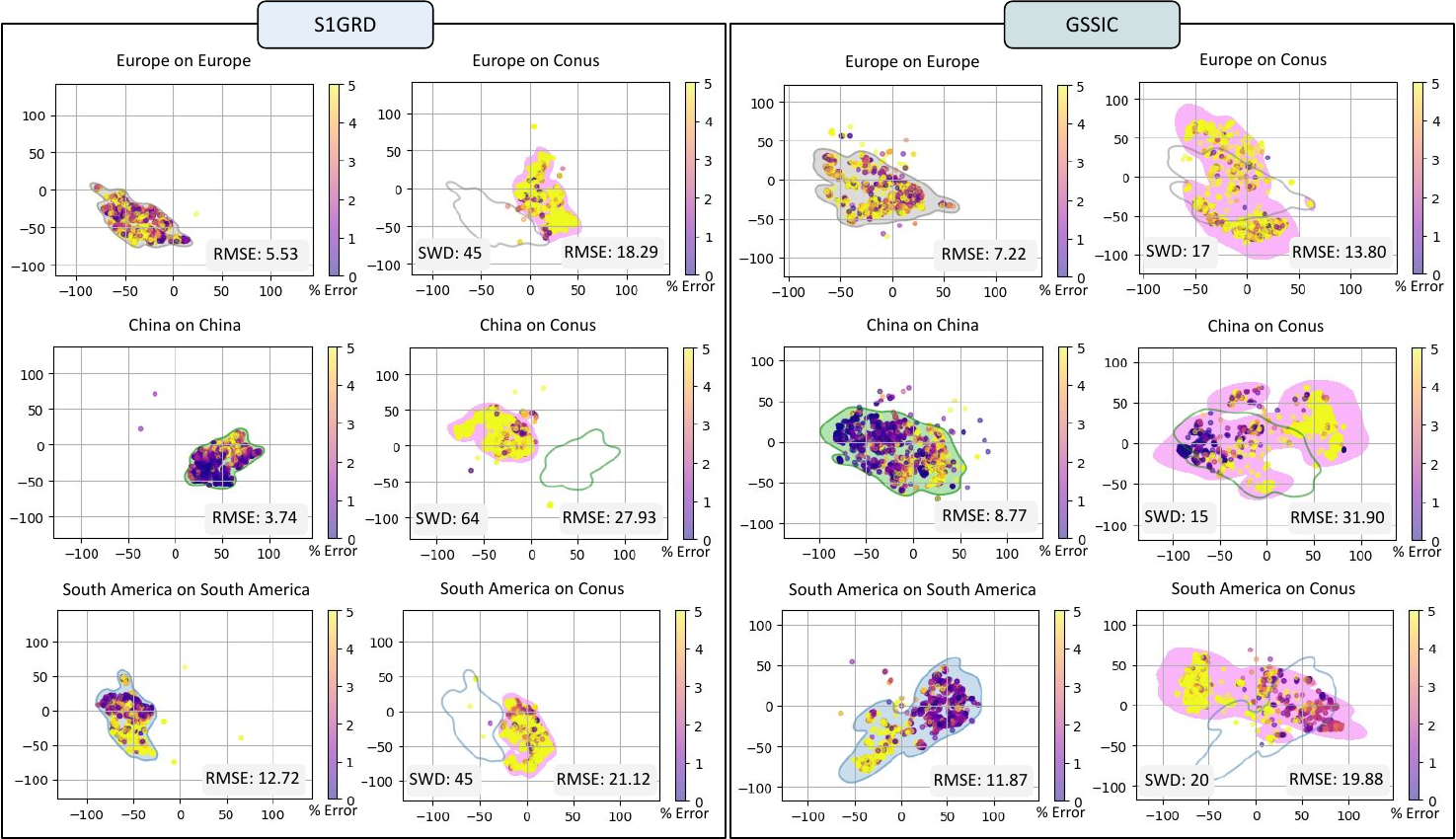}
    \caption{Embedding space and prediction errors (in \%) for models pre-trained on S1GRD or GSSIC. The first (third) column shows the embeddings and prediction error for 2000 example tiles in the respective fine-tuning areas for S1GRD (GSSIC), while the second (fourth) column shows the embeddings and prediction errors when performing inference of the fine-tuned models on 2000 tiles in Conus. The Sliced Wasserstein Distance (SWD) of the embedding distributions and the root mean square error (RMSE[\%]) for inference on the entire test is included.}
    \label{fig:embedding_conus}
\end{figure}

\begin{figure}[ht]
    \centering
    \includegraphics[width=\linewidth]{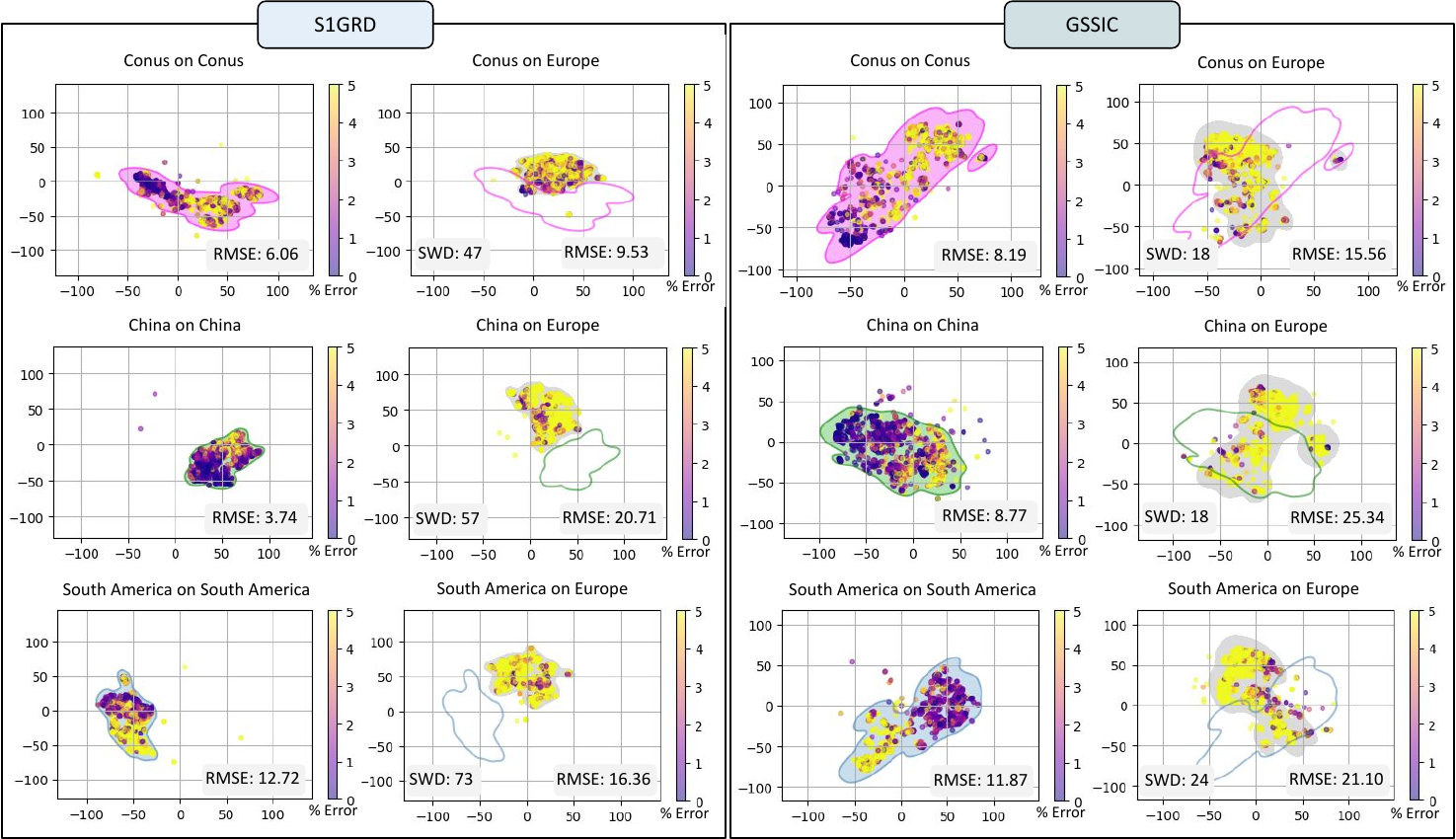}
    \caption{Embedding space and prediction errors (in \%) for models pre-trained on S1GRD or GSSIC. The first (third) column shows the embeddings and prediction error for 2000 example tiles in the respective fine-tuning areas for S1GRD (GSSIC), while the second (fourth) column shows the embeddings and prediction errors when performing inference of the fine-tuned models on 2000 tiles in Europe. The Sliced Wasserstein Distance (SWD) of the embedding distributions and the root mean square error (RMSE[\%]) for inference on the entire test is included.}
    \label{fig:embedding_europe}
\end{figure}

\begin{figure}[ht]
    \centering
    \includegraphics[width=\linewidth]{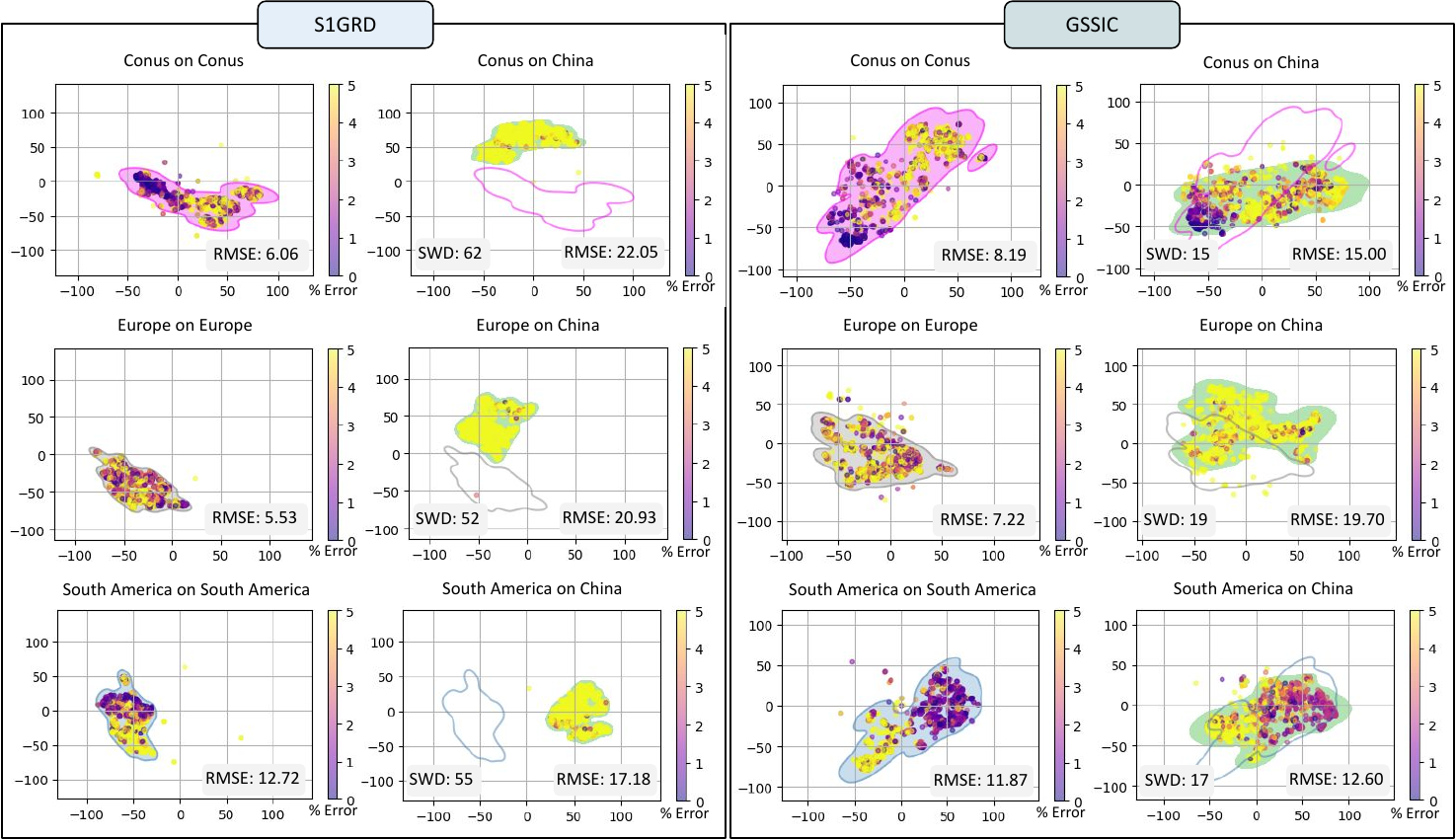}
    \caption{Embedding space and prediction errors (in \%) for models pre-trained on S1GRD or GSSIC. The first (third) column shows the embeddings and prediction error for 2000 example tiles in the respective fine-tuning areas for S1GRD (GSSIC), while the second (fourth) column shows the embeddings and prediction errors when performing inference of the fine-tuned models on 2000 tiles in China. The Sliced Wasserstein Distance (SWD) of the embedding distributions and the root mean square error (RMSE[\%]) for inference on the entire test is included.}
    \label{fig:embedding_china}
\end{figure}
}

\end{document}